\documentclass[10pt,twocolumn,letterpaper]{article}

\usepackage{iccv}
\usepackage{times}
\usepackage{epsfig}
\usepackage{graphicx}
\usepackage{amsmath}
\usepackage{amssymb}
\usepackage[colorlinks,linkcolor=blue]{hyperref}

\iccvfinalcopy 

\def\iccvPaperID{3239} 

\ificcvfinal\fi
\begin{document}

\title{A Neural Network for Detailed Human Depth Estimation from a Single Image}


\author{Sicong Tang$^{1,*}$ \ \ Feitong Tan$^{1,}$\thanks{These authors contributed equally to this work.} \ \ Kelvin Cheng$^{1}$ \ \ Zhaoyang Li$^{1}$ \ \ Siyu Zhu$^{2}$ \ \ Ping Tan$^{1}$ \\
$^{1}$ Simon Fraser University \ \ $^{2}$ Alibaba A.I Labs  \\
{\tt\small \{sta105, feitongt, kelvinz, zla143, pingtan\}@sfu.ca,  siting.zsy@alibaba-inc.com
}}


\maketitle

\begin{abstract}
This paper presents a neural network to estimate a detailed depth map of the foreground human in a single RGB image. The result captures geometry details such as cloth wrinkles, which are important in visualization applications. To achieve this goal, we separate the depth map into a smooth base shape and a residual detail shape and design a network with two branches to regress them respectively.
We design a training strategy to ensure both base and detail shapes can be faithfully learned by the corresponding network branches. Furthermore, we introduce a novel network layer to fuse a rough depth map and surface normals to further improve the final result.
Quantitative comparison with fused `ground truth' captured by real depth cameras and qualitative examples on unconstrained Internet images demonstrate the strength of the proposed method. Our code will be released at \href{https://github.com/sfu-gruvi-3dv/deep_human}{Link} 
\end{abstract}

\section{Introduction} 
Understanding human images is an important problem in computer vision with many applications ranging from human-computer-interaction and surveillance to telecommunication. Many works~\cite{hourglass_2016,vnect_2017,Toshev_2014_CVPR,sun2018integral,joints_baseline_2017,max_margin_joint3d_2015} have been developed to recover 2D or 3D skeleton joints from a RGB image. 
Since the skeleton only captures sparse information of the human body, DensePose~\cite{alp2018densepose} estimates a dense UV map (i.e. a correspondence map between the input image and a 3D template model). But this UV map can not recover 3D shape without additional 3D pose information, which limits its application.

On the other hand, there are many works~\cite{scape2005,smpl2015,smplify2016,smpl_est_2016,smpl_est_2017,up_2017,scape_est_2007,scape_est_2009, omran2018neural} to recover a dense 3D deformable  model of the human body from a single image, e.g. the SCAPE~\cite{scape2005} and SMPL~\cite{smpl2015} models, which are learned from a large dataset of scanned body shapes. While generating 3D models, these methods only inference the naked body shape without capturing the clothes details. 

This paper aims at recovering a detailed depth map for the foreground human object from a single RGB image. This problem has been studied in the earlier work~\cite{varol17_surreal} with synthetic human images.  Another recent work~\cite{bodynet2018} recovers a volumetric 3D model of the imaged person. Results from both methods are too coarse for many applications. In comparison, we design a neural network to estimate highly detailed depth maps that are fine enough to capture cloth wrinkles, which might potentially be exploited for telepresence applications like the Microsoft Holoportation~\cite{fusion4d_2016}.

Our network is designed with two novel insights. Firstly, we argue it is important to separate the depth to a smooth base shape and a residual detail shape and regress them respectively. The base shape captures the large overall geometry layout, while the detail shape captures small bumps such as cloth wrinkles. The value range of the base shape is at the scale of one meter, while that of the detail shape is at a few centimeters. Thus, we design a network with two branches for the base and detail shapes respectively to facilitate the training process. 
Specifically, we propose a 2-stage training strategy to ensure the effectiveness of this separation.
These two branches are trained respectively in the 1st stage and then finetuned together in the 2nd stage.
Secondly, we follow the intuition in~\cite{Zhang_2018_CVPR} to estimate surface normals to facilitate depth map estimation. Specifically, we generalize the algorithm in~\cite{Nehab:2005:ECP} that fuses surface normals and a coarse depth to an iterative formulation. In this way, we build a parameter-free network layer to fuse the estimated normals and a coarse depth map for improved results.

Our final network captures visually appealing detailed depth images from a single RGB image. The evaluation on our own captured real data and some unconstrained online images demonstrate its effectiveness.  We will publish our dataset and source code with the paper to facilitate further research.
\section{Related works}

\textbf{3D Human Pose Estimation.} With the recent development of deep convolutional neural networks (CNNs), there are significant improvements on 3D human pose estimation~\cite{joints_baseline_2017,sun2018integral,max_margin_joint3d_2015, vnect_2017}. Despite the differences in network architectures, many works~\cite{hourglass_2016,volumetric_joint3d_2017,vnect_2017,sun2018integral,bodynet2018} use a likelihood heatmap to represent the distribution of each joint's location and show better performance than directly regressing the joint location. Instead of taking the maximum from a heatmap, Sun \textit{et al.} \cite{sun2018integral} compute the expected coordinates from a heatmap to reduce the artifacts due to quantization. 
The recent work DensePose\cite{alp2018densepose} is even able to recover dense UV coordinate for each pixel on human body. 
Unlike our method, most of these methods only recover sparse 3D joint positions.
While DensePose provides dense result, its result is not in 3D but rather a 2D UV coordinate map.
We adopt a pose estimation network as an intermediate layer and use its results to guide the dense depth recovery.


\textbf{Body Shape Estimation.} The 3D shape of a human body can be parameterized by the SCAPE or SMPL models~\cite{scape2005,smpl2015} with two sets of independent parameters, controlling the skeleton pose and body shape respectively. Both models are derived from a large set of scanned 3D human shapes. Given these parametric human models, many methods~\cite{scape2005,smpl2015,smplify2016,smpl_est_2016,smpl_est_2017,up_2017, omran2018neural}  recover dense human body shape from a single RGB image by estimating the shape and pose parameters. Meanwhile, there are also some non-paramterized methods~\cite{varol17_surreal,bodynet2018} which directly regress discretized body shape representation from a RGB image. The above methods only recover the 3D shape of the naked human body and geometry details like the clothes are not modeled, which make them not suitable for visualization tasks. While the method~\cite{alldieck2018video} can predict the SMPL model with clothe wrinkle, it needs to be fed a video of a moving person with designed pose. To overcome this limitation, our network aims at recovering shape details from a single image.


\textbf{Generic Dense Depth Estimation.} Depth estimation from a single image has gained increasing attention in the computer vision community. Most works like \cite{wang2018learning,yin2018geonet,mahjourian2018unsupervised,kumar2018depthnet,zhan2018unsupervised,unsupervised_ego_2017, eigen2014depth, laina2016deeper} are proposed for indoor and outdoor scenes.
We focus on depth estimation of humans, which allows us to build much stronger shape prior than these generic depth estimation methods. Specifically, our network first estimates the skeleton joints and a body part segmentation to facilitate the depth estimation.


\section{Overview}
\begin{figure*}
\begin{center}
   \includegraphics[width=0.90\linewidth]{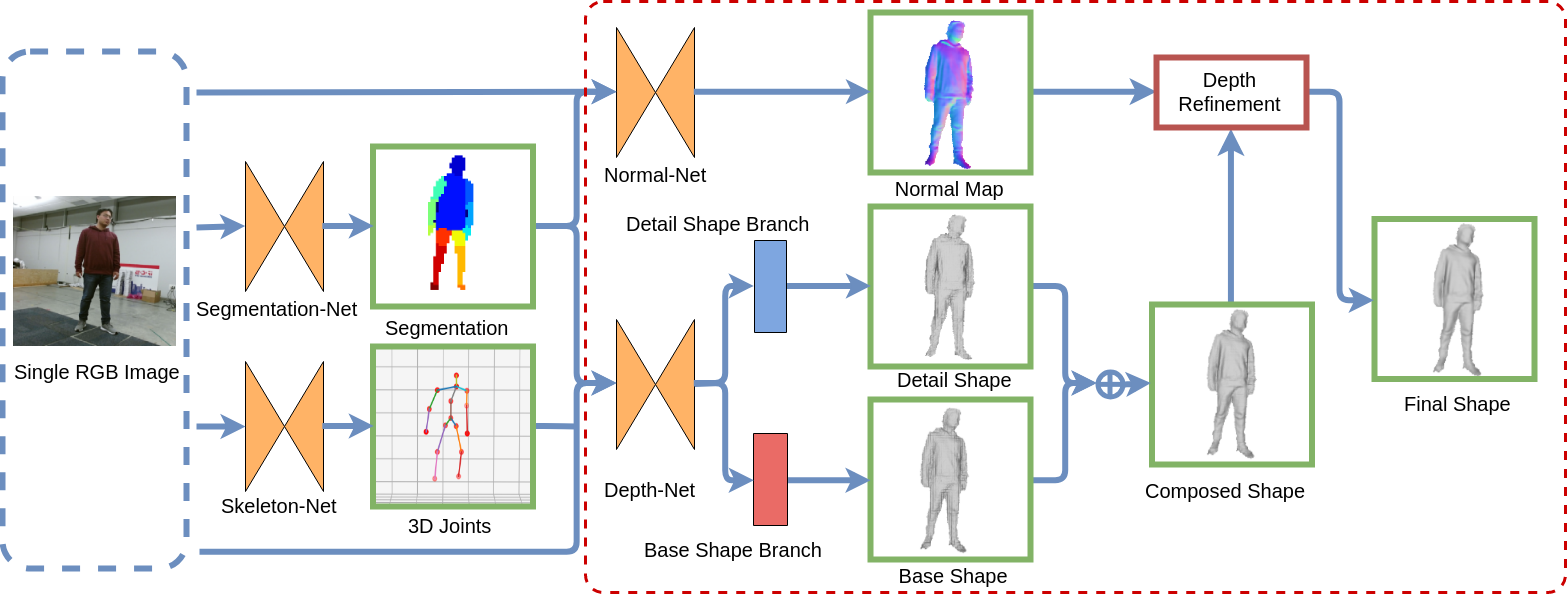}
\end{center}
\vspace{-2mm}
\caption{The structure of our proposed network. The Skeleton-Net and Segmentation-Net generate the heatmaps of 3D skeleton joints and body part segmentation respectively. Their results are further fused with the input image to compute the base shape and detail shape via the Depth-Net. In a separate branch, the Normal-Net estimates a surface normal map. The composed shape and normal map are further fused in the depth refinement module to produce the final result.}
\label{fig:pipeline}
\vspace{-2mm}
\end{figure*}

The overall structure of the proposed network is shown in Figure~\ref{fig:pipeline}. The input is a 256$\times$256 3-channel RGB image containing a human as the foreground. The network first computes the heatmaps of the 3D skeleton joints and a body part segmentation through two Hourglass networks~\cite{hourglass_2016}, which are referred as \emph{Skeleton-Net} and \emph{Segmentation-Net} respectively in this paper. We then concatenate the outputs of these two modules with the input RGB image and feed them to the \emph{Depth-Net} to compute the initial depth maps, which consists of a base shape and a detail shape. 

In a separate branch, another Hourglass network, referred as \emph{Normal-Net}, computes a surface normal map of the human body from the input RGB image and the segmentation mask generated by the Segmentation-Net.
We then compose the base shape and detail shape, and fuse the composed shape  and normal map through a parameter-free shape refinement module to produce the final shape.

During training, we first pre-train the Skeleton-Net, Segmentation-Net, and Depth-Net on synthetic data~\cite{varol17_surreal} respectively. 
Meanwhile, the Normal-Net is pre-trained on the deforming fibre dataset~\cite{bednarik2018learning}. Then we finetune the complete network on the real image dataset captured by ourselves with a depth camera, while keeping the parameters of Skeleton-Net and Segmentation-Net fixed.

\section{Segmentation and Skeleton Networks}


Inspired by the BodyNet~\cite{bodynet2018}, 3D joints and body part segmentation are highly correlated with the final estimation of human shapes.
We therefore apply two Hourglass networks~\cite{hourglass_2016} to estimate the heatmaps of 3D joints and a body part segmentation from the input RGB image.
As demonstrated in the ablation studies, this intermediate supervision of 3D joints and body part segmentation is essential for the depth estimation, especially for the base shape.

Here, a human body contains 16 joints and 14 body parts.
For each joint, our Skeleton-Net predicts a heatmap indicating the probability of its position~\cite{pavlakos2017coarse}.
The 3D joints are defined in the camera coordinate system, where the $xy$-axes are aligned with the image axes, and the $z$ axis is the camera principal direction. We discretize the $z$ coordinate between $[-0.6, 0.6]$ meters into 19 bins and set the depth of the pelvis joint as $0$.
The $x$ and $y$ coordinates are discretized into 64 bins over the image plane. 
Therefore, the network estimate a heatmap of size 64$\times$64$\times$19 for each joint, resulting in a skeleton representation as a 64$\times$64$\times$19$\times$16 heatmap.

Unlike \cite{bodynet2018}, we discard the 2D joint estimation sub-network and predict the 3D joints directly, which makes our network more compact. In order to achieve good accuracy with this compact network, we adopt the integral regression~\cite{sun2018integral} to train the Skeleton-Net.


For body part segmentation, the Segmentation-Net predicts the probability heatmap for the 14 body parts and the background, which results in a 64$\times$64$\times$15 heatmap. Following the previous work of human part segmentation~\cite{oliveira2016deep}, we adopt the spatial cross-entropy loss in training.

\section{Depth Estimation Network}\label{sec:roughdepth}
To better estimate a detailed depth map with cloth wrinkles, we divide the depth map of a human body into a smooth base shape and a residual detail shape: the base shape captures the main geometry layout of the human body, while the detail shape is responsible for describing local geometry details such as cloth wrinkles.

\begin{figure}
\includegraphics[width=1\linewidth]{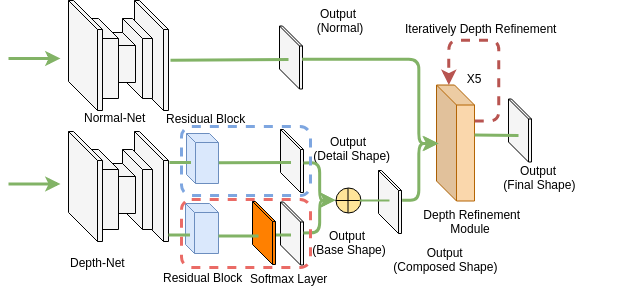}
\caption{Architecture of Depth-Net together with base shape branch and detail shape branch, Normal-Net and Depth Refinement Module.
The branches in blue and red dashed rectangles correspond to the detail and base shape branch respectively.}
\label{fig:rough_depth_net}
\end{figure}

As shown in Figure~\ref{fig:rough_depth_net} which corresponds to the part in the red dashed rectangle in Figure~\ref{fig:pipeline}, the Depth-Net is composed of a U-Net~\cite{ronneberger2015u} and a two-branch architecture. The concatenation of the RGB image and bilinearly-upsampled heatmaps (64$\times$64 to 256$\times$256) of 3D joints and segmentation is fed into this network, and the two branches, namely base and detail shape branch, output a base shape and detail shape separately. 
Because the human layout is approximately one-meter range with low frequency in image plane and the detailed cloth wrinkles is just several centimeters with higher frequency, the two branches concentrate on these two different distributions respectively.

To effectively train the Depth-Net, we set the median of the ground-truth depth as 0 and decouple this zero-median depth image into a base shape and detail shape. 
Specifically, we apply the bilateral filter to the depth image to smooth out the details and obtain the base shape.
We denote this base shape as $F(D_{gt})$, where $D_{gt}$ is the ground-truth depth image and $F(\cdot)$ is the operation of the bilateral filter. In our work, the depth sigma is set as $0.10$ meters and the space sigma is set as $75$ pixels for the bilateral filter.
The ground-truth of the detail shape is computed as a residual $R_{gt}$:
\begin{equation}
R_{gt} = D_{gt} - F(D_{gt}).
\end{equation}

For the base shape, we discretize the depth range between [-0.6, 0.6] meters into 19 bins for each pixel. The softmax layer which follows a residual block in the base branch generates a 256$\times$256$\times$19 heatmap indicating the probability of the depth bin. Afterwards, a 256$\times$256 depth map can be calculated from the heatmap by an integral operation~\cite{sun2018integral}. Meanwhile, in the detail branch, a residual depth map of detail shape which has a higher frequency is regressed directly.  At last, we add the base shape and detail shape together to obtain the composed shape.

In order to guide the base and detail branch to focus on their target domain (base shape and detail shape) , we train our Depth-Net following a two-stage strategy.
In the first stage, the base and detail branch are pre-trained separately to obtain well-conditioned initial values.
In the second stage, we perform end-to-end training on three combined weights with the supervision of the intermediate base and detail shape branches.

\subsection{Training stage 1}

Once we have the ground-truth base and detail shape, we pre-train these two branches independently with the following loss functions:
\begin{align} \label{equ:separate_depth_loss}
\begin{split}
L_{base} &= H( D_{base} - F(D_{gt}) , \alpha_1 ), \\
L_{detail} &= H( D_{detail} - R_{gt} , \alpha_2),
\end{split}
\end{align}
where $D_{base}$ and $D_{detail}$ are the base and detail depth to be regressed respectively.
$H(x, \alpha)$ is the Huber loss function, $\alpha_1$ and $\alpha_2$ are set as 0.2 meters and 0.05 meters.
Here, $H(x, \alpha)$ is defined as:
\begin{equation} \label{equ:HuberL}
    H(x, \alpha) = \left\{ \begin{array}{rcl}
       0.5 x^2, &  x \leq \alpha,\\
      0.2 (|x|-\alpha), &  x > \alpha.
    \end{array} \right.
\end{equation}


This pre-training helps the two branches focus on different aspect of the shape estimation, where the base shape captures the main geometry layout and the detail shape adds on high-frequency wrinkles. 


\subsection{Training stage 2}

In this stage, we jointly train these two branches by using the combined loss $L$ below:
\begin{align} \label{equ:joint_depth_loss}
\begin{split}
L &= \beta_1 L_{base} + \beta_2 L_{detail} + \beta_3 L_{composed},
\end{split}
\end{align}
where $\beta_1$, $\beta_2$, $\beta_3$ are set as $1,1,15$. Here, the composed loss $L_{composed}$ is formulated as:
\begin{align} \label{equ:joint_depth_loss}
\begin{split}
L_{composed} &= T( D_{base} + D_{detail} - D_{gt}, \alpha_3),
\end{split}
\end{align}
where $\alpha_3$ is set to $0.05$ meters in our experiments. $T(x,\alpha)$ is the truncated $L_1$ loss and it is defined as:
\begin{equation} \label{equ:truncatedL1}
    T(x, \alpha) = \left\{ \begin{array}{rcl}
        x, &  x \leq \alpha,\\
        \alpha, &   x > \alpha.
    \end{array} \right.
\end{equation}

The stage 2 improves the consistency between the combined shape and the ground truth, and the truncated $L_1$ loss is used to define the composed loss $L_{composed}$ which clips the loss value to a bounded range. This truncated loss helps to avoid the training being biased by large shape errors due to imprecise poses, which could overwhelm the errors due to missing cloth wrinkles. As we will see in experiments, this loss helps the detail shape branch to capture details. 



\section{Normal Network and Depth Refinement}
As observed in~\cite{Zhang_2018_CVPR}, regressing surface normal is often more reliable than regressing depth directly.
We include a network to regress the surface normal at every pixel and use this information to refine the composed depth.

\subsection{Normal Network}
Here, a Hourglass network takes a RGB image concatenated with a segmentation mask from the Segmentation-Net as input and outputs a normal map.

This network is trained with the ground-truth normal computed from the ground-truth depth map $D_{gt}$.
To compute the ground-truth normal $N_{gt}$, we take the nearby 3D points at each pixel to estimate its normal direction by the standard linear least square fitting. The loss function is the mean angular difference between the ground-truth and the regressed normal.

\subsection{Depth Refinement}
We fuse the composed depth and surface normal here to improve the depth quality. Similar to~\cite{nehab2005efficiently}, we formulate the problem with two constraints. Firstly, the tangent vector of the final shape should be perpendicular to the input surface normal at each pixel. Secondly, the final shape should be close to the initial shape.
Rather than solving a large linear system for a global optimization which is impractical for a neural network, we introduce an iterative solution.  

At each iteration, we update the depth assuming its neighboring depth is fixed. Concretely, we define $(N_{ix}, N_{iy}, N_{iz})$ as the normal of pixel $i$ in $x,y,z$ directions, and $(X_i^n, Y_i^n, Z_i^n)$ as the position of pixel $i$ after the $n$-th iteration.
At the $n$+$1$-th iteration, we update $Z_i^{n+1}$ for each pixel $i$ with the depth of neighboring pixels fixed at $Z_j^n$. Here, $j \in \mathcal{N}_i$ is a neighboring pixel of $i$ and there are 4 neighbors for each pixel in cardinal directions.
The update function is defined as: 
\begin{equation}
    Z_i^{n+1} = \lambda Z_i^{0} + (1-\lambda) \frac{\sum_{j \in \mathcal{N}_i} \left( Z_{ij}^n + Z_{ji}^n \right) }{8},
\end{equation}
where $Z_{ij}^n$ is the depth of $i$ that makes the edge $ij$ and $N_j$ perpendicular, and $Z_{ji}^n$ is the depth of $i$ that makes $ij$ and $N_i$ perpendicular. Specifically, they can be computed as:
\begin{equation}
    \begin{aligned}
        Z_{ij}^{n} &= \frac{N_{jx}(X_j^n - X_i^n) + N_{jy}(Y_j^n - Y_i^n) + N_{jz}Z_j^n}{N_{jz}}, \\
        Z_{ji}^{n} &= \frac{N_{ix}(X_j^n - X_i^n) + N_{iy}(Y_j^n - Y_i^n) + N_{iz}Z_i^n}{N_{iz}}.
    \end{aligned}
\end{equation}
Here, $\lambda$ is the hyper-parameter (fixed at 0.4).

\begin{figure}
\center
\includegraphics[width =0.95 \linewidth]{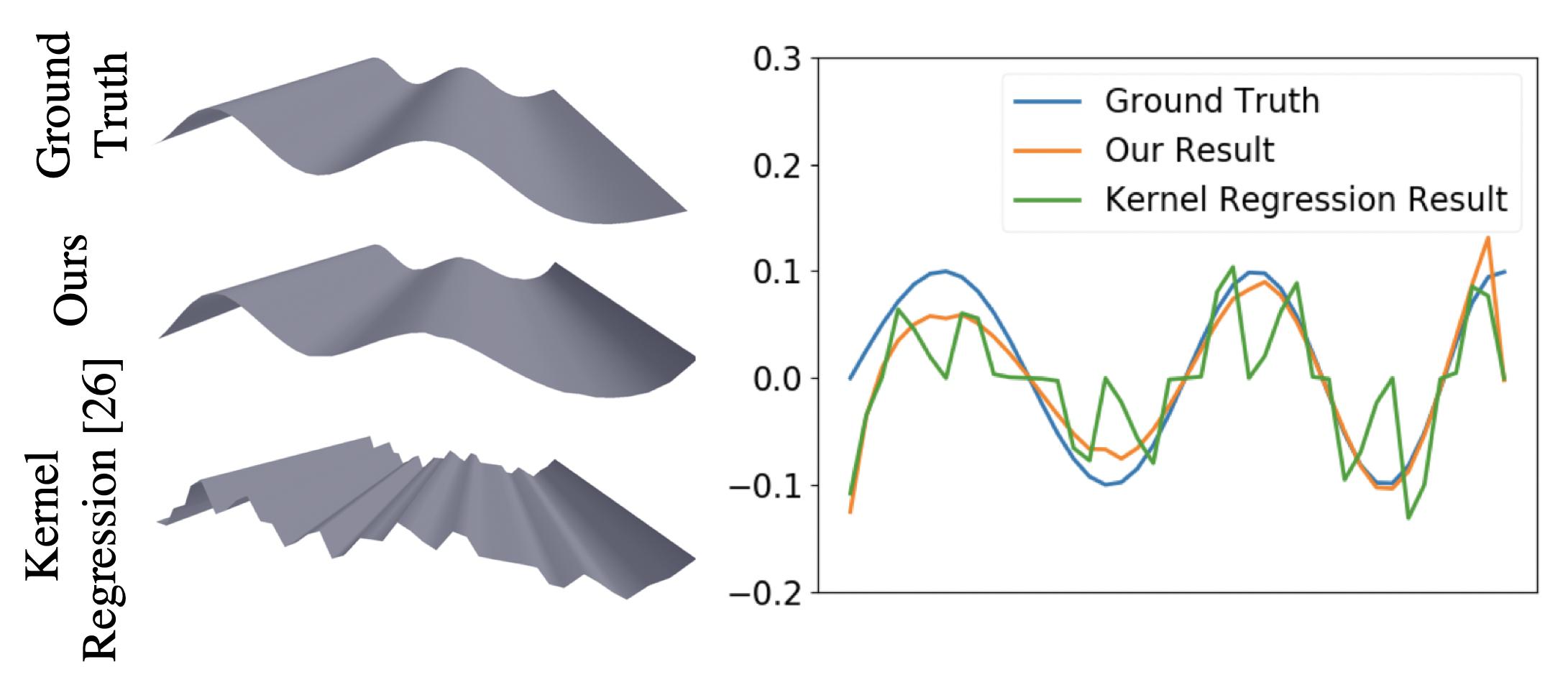} 
\caption{Comparison of our depth refinement with~\cite{qi2018geonet} on a toy example of Sine curve. Left: ground-truth, results from our method and the~\cite{qi2018geonet} (from top to bottom). Right: sectional view of these results.}
\label{fig:3d_sine_compare}
\end{figure}

\begin{figure}
\center
\includegraphics[width =0.80 \linewidth]{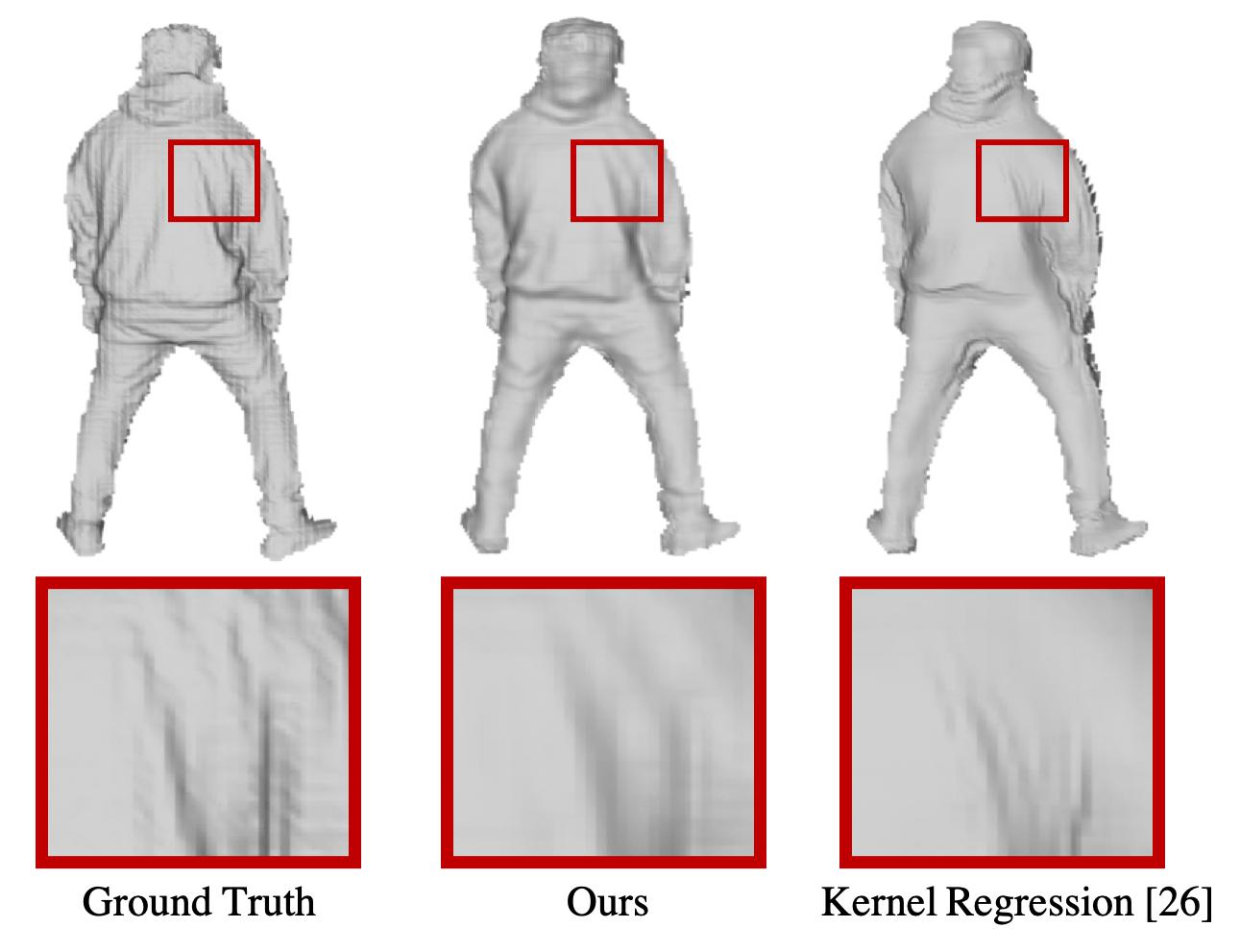}
\caption{Comparison of our depth refinement with the `Kernel Regression' in~\cite{qi2018geonet} on a real data. From left to right, there are the ground truth shape, results of our method and the `Kernel Regression' respectively.
}
\label{fig:real_compare}
\end{figure}

The above shape refinement is iterated for 5 times in our network to simulate the iterative solution of the original energy equation in~\cite{nehab2005efficiently}.
Figure~\ref{fig:3d_sine_compare} compares our method with the `Kernel Regression' layer~\cite{qi2018geonet} on a toy example, which is also designed to fuse the surface normal and depth.
Figure~\ref{fig:real_compare} shows a comparison with the work~\cite{nehab2005efficiently} on real data and our method also produces more convincing result.

\begin{figure*}[h!]
\center
 \includegraphics[width= 0.95 \linewidth]{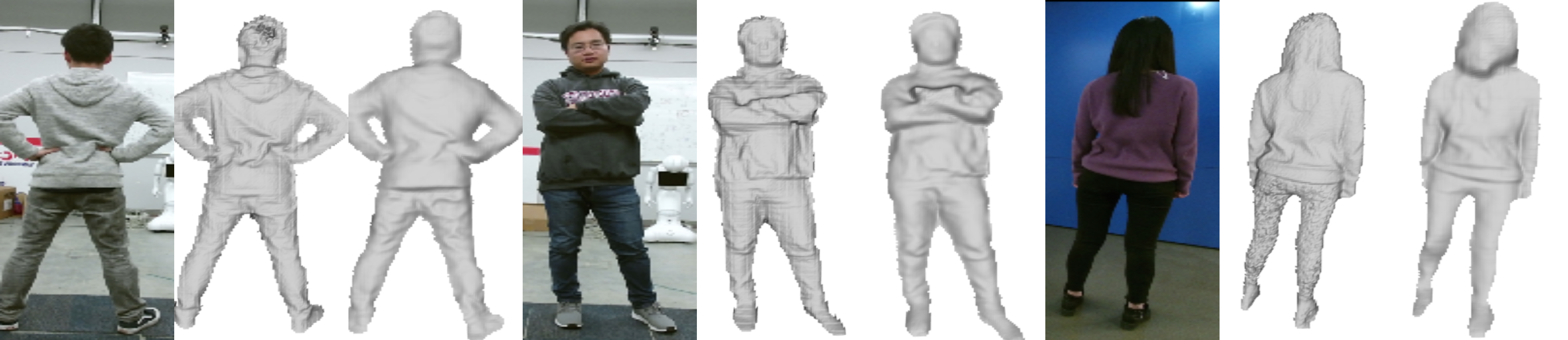}\\
 \includegraphics[width= 0.95 \linewidth]{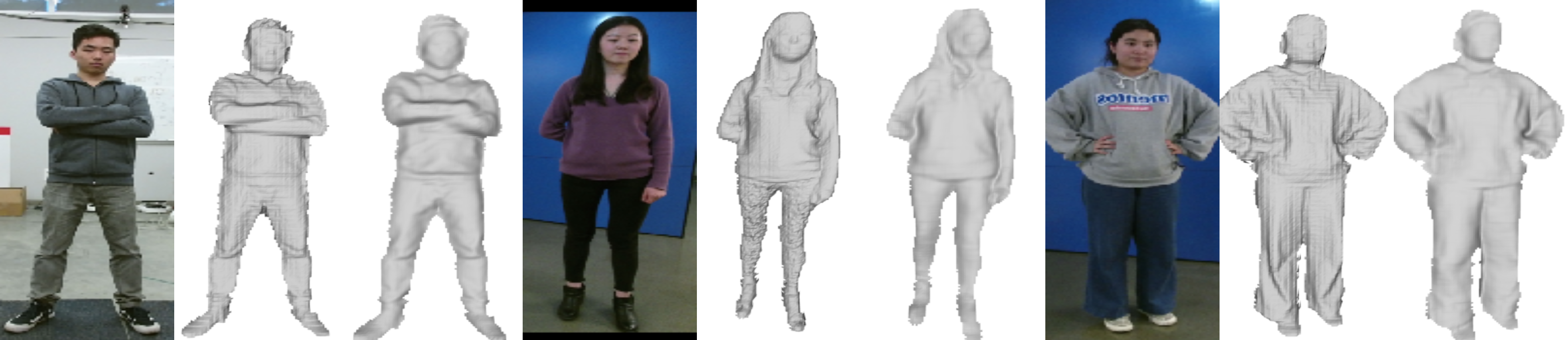}
\caption{Some results on the testing data. From left to right, these images are: the single input RGB image, the ground truth shape and our result. It can be seen that our method is able to recover the main layout as well as certain geometry details. Note that our results are trained on the noisy raw depth images captured by the Kinect2 camera, however, our network is still able to give polished results.}
\label{fig:comparison_pipeline}
    \vspace{-4mm}
\end{figure*}

\section{Experiment}
To demonstrate the effectiveness of our method, we evaluate it using ablation studies and both qualitative and quantitative comparisons with other relevant works~\cite{varol17_surreal,bodynet2018}, a surface-from-normals method~\cite{kovesi2005shapelets} and a general depth estimation network~\cite{laina2016deeper}. To test the performance of human shape estimation with fine-grain geometrical details, we build up our own dataset for evaluation.

\textbf{Implementation Details.} All input RGB images are cropped to center the person with size 256$\times$256, assuming that the bounding box of person is given. The RMSprop~\cite{rmsprop} algorithm with a fixed learning rate of 1$\times10^{-5}$ is used. We first train our Segmentation-Net, Skeleton-Net and Depth-Net on SURREAL~\cite{varol17_surreal}, a large-scale synthetic human body dataset without geometrical details.
At this stage, the batch size is set to 6 for these three networks, and for Depth-Net we only add base shape loss to train base shape branch since the synthetic data does not have much geometrical details. The Normal-Net is pre-trained on a deforming fibre dataset~\cite{bednarik2018learning}. 
After the base shape branch of Depth-Net converges, which takes 10 epochs, 12 hours on a GTX 2080 GPU, we fix the weight of Skeleton-Net and Segmentation-Net and fine-tune the Depth-Net and Normal-Net jointly on our own captured data with a batch size of 1. It takes another 12 epochs, 10 hours for stage 1 and another 8 epochs, 6 hours on stage 2. During inference, our network takes 75.5ms for the whole pipeline, and 61.1ms without iterative depth refinement on a RTX 2080.

\textbf{Dataset.}\
We collect a RGBD dataset for real persons. Here the dataset contains 26 different people performing simple actions captured by a Microsoft Kinect2 camera. 

For the training data, we capture approximate 800 frames for each person, leading to over 20,000 training depth images in total. For quantitative evaluation, we use depth cameras to capture video clips of a person with a fixed pose and employ the InfiniTAM~\cite{InfiniTAM_ISMAR_2015} to fuse captured sequences. The high-quality depth maps are rendered according to the fused mesh and camera poses with Blender~\cite{blender}. Our testing data contains 5 different persons, each person is captured with 12 different poses and 3 different clothing styles. 

Note that we only use the fused depth maps for evaluation,  the training data are raw depth maps since it is infeasible to fuse all the meshes with thousands of poses for rendering the depth maps.


\begin{table}[]
\begin{tabular}{|l|lll|l|}
\hline
\multicolumn{1}{|c|}{Methods} & \multicolumn{3}{|c|}{Accuracy}                                                                           & MAE \\ \cline{2-4}
\multicolumn{1}{|c|}{}                        & \multicolumn{1}{c}{1.25cm}           & \multicolumn{1}{c}{2.5cm}           & \multicolumn{1}{c|}{5.0cm} &                      \\ \hline
Ours (Final Shape)                             & \multicolumn{1}{l|}{\textbf{30.06}} & \multicolumn{1}{l|}{\textbf{51.57}} & 75.76             & 3.208       \\ \hline
Ours (Base + Detail)                           & \multicolumn{1}{l|}{29.24}          & \multicolumn{1}{l|}{50.93}          & 75.52                      & 3.282                \\ \hline
Ours (Base Shape)                              & \multicolumn{1}{l|}{28.03}          & \multicolumn{1}{l|}{50.10}          & 75.32                      & 3.396                \\ \hline
Ours (Off-the-Shelf)                                & \multicolumn{1}{l|}{28.57}          & \multicolumn{1}{l|}{50.70}          & \textbf{76.54}                      & 3.546                \\ \hline
SURREAL \cite{varol17_surreal}                                      & \multicolumn{1}{l|}{21.32}          & \multicolumn{1}{l|}{37.52}          & 50.06                      & 3.976                \\ \hline
BodyNet \cite{bodynet2018}                                      & \multicolumn{1}{l|}{17.14}          & \multicolumn{1}{l|}{32.59}          & 56.98                      & 4.366                \\ \hline
Laina et al. \cite{laina2016deeper}                                  & \multicolumn{1}{l|}{19.84}          & \multicolumn{1}{l|}{36.48}          & 60.94                      & 4.902                \\ \hline
Kovesi et al. \cite{kovesi2005shapelets}                               & \multicolumn{1}{l|}{15.51}          & \multicolumn{1}{l|}{29.87}          & 55.39                      & 5.789                \\ \hline
\end{tabular}
\\[1pt]
\caption{Performance of depth estimation on the test set. `Ours (Base)' stands for the base shape without adding detail wrinkles. `Ours (Base + Detail)' refers to the composed shape before the depth refinement.}
\label{tab:AUC}
\vspace{-2mm}
\end{table}

\begin{figure}
\includegraphics[width=1.0\linewidth]{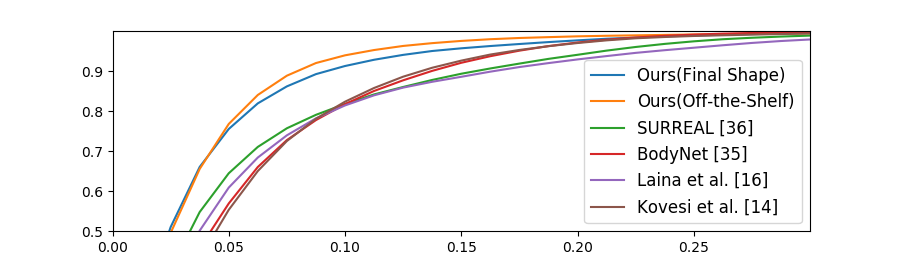}
\caption{Cumulative Distribution Function of depth error of our method and comparison methods~\cite{varol17_surreal,bodynet2018,laina2016deeper}. }
\label{fig:CDF}
\vspace{-2mm}
\end{figure}

\subsection{Quantitative Results}
Figure~\ref{fig:comparison_pipeline} shows our results compared with the fused ground-truth depth. We can see that our method can successfully capture cloth wrinkles and produce visually appealing 3D mesh from testing real images, despite our model is trained on the noisy raw depth images. 



\textbf{Comparison with ~\cite{varol17_surreal,laina2016deeper,bodynet2018,kovesi2005shapelets}.}
There are only a few works that can compute a depth map of human body from a single image. We compare with the two most recent works~\cite{varol17_surreal,bodynet2018} and a representative general depth estimation framework~\cite{laina2016deeper}, and since we use normal map to refine human depth in our framework, we also evaluate a surface-from-normals method~\cite{kovesi2005shapelets} with the normals from our Normal-Net. At last, to show the generalizability of our network, we replace our segmentation and 3D pose estimation module with off-the-shelf networks \cite{ronneberger2015u,varol17_surreal} and evaluate the performance of Depth-Net. To make the comparison fair, we fine-tune~\cite{varol17_surreal,laina2016deeper} on our dataset. Unfortunately, the  BodyNet~\cite{bodynet2018} needs a volumetric shape representation and its loss function contains the multiview constraints, thus it can not be fine-tuned on our data.
Here, the pixel accuracy as percentage of pixels with depth errors smaller than some specified threshold is employed as the evaluation metric. It shows in Table~\ref{tab:AUC} that the final shape after refinement always produces the highest accuracy. Here we notice that our network still works well with off-the-shelf segmentation and 3D pose estimation methods, and deducing the correct human shape just from normal is difficult due to noisy normal estimation and depth discontinuities. We also use the Mean Absolute Error (MAE) as a more global metric to prove that our method captures not only details but also overall shapes. Furthermore, we plot the  Cumulative Distribution Function (CDF) of the shape errors by different methods in Figure~\ref{fig:CDF}, which illustrates that our method outperforms others with different shape scales.

\begin{figure}
\begin{center}
\includegraphics[width=1.0\linewidth]{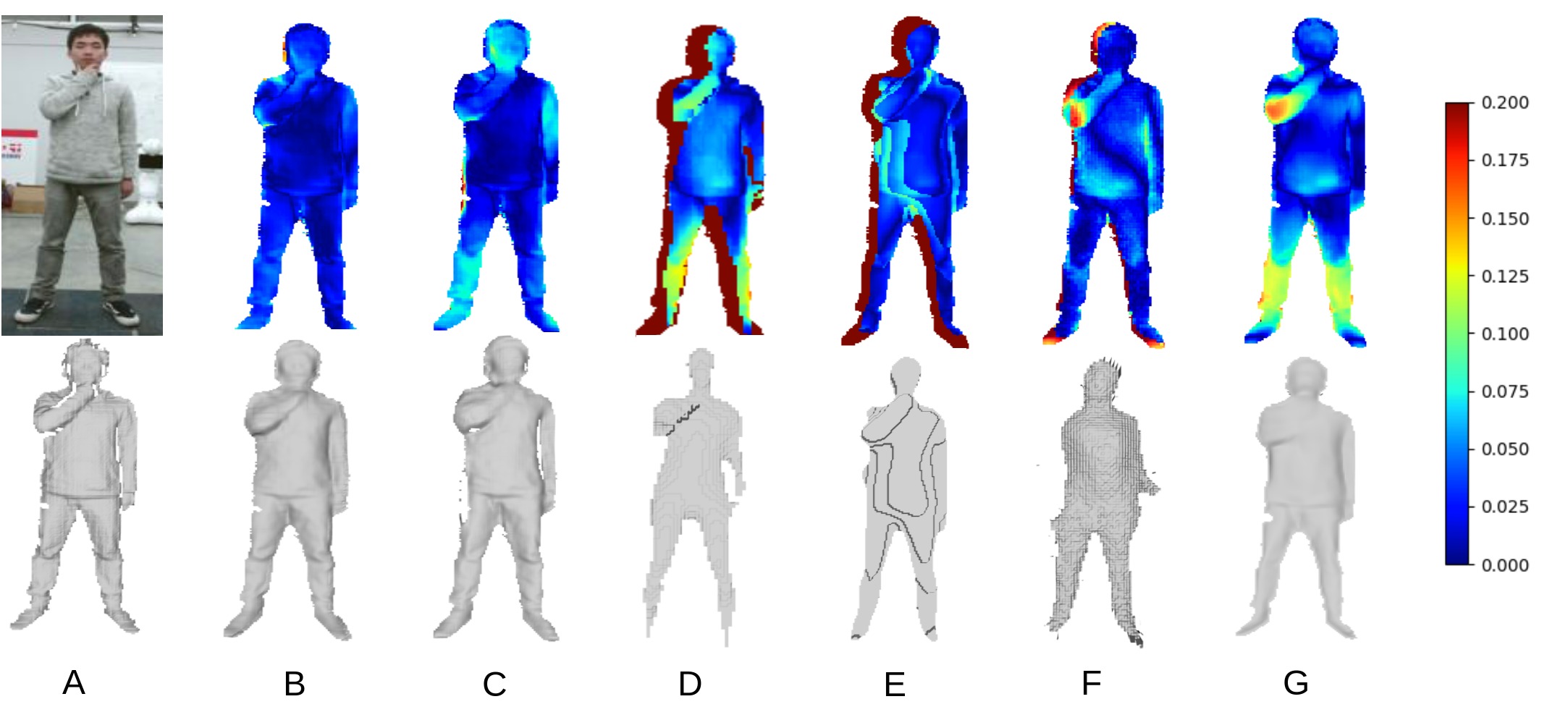}
\caption{Qualitative comparisons. The first row shows the heatmaps for depth errors, while the second row shows the recovered mesh. Left to right columns: A. Ground truth, B. Ours (Final Shape), C. Ours (Off-the-Shelf),  D. SURREAL~\cite{varol17_surreal}, E. BodyNet~\cite{bodynet2018}, F. Laina et al.~\cite{laina2016deeper} and G. Kovesi et al.~\cite{kovesi2005shapelets} respectively.}
\label{fig:comparison}
\vspace{-2mm}
\end{center}
\end{figure}

Figure~\ref{fig:comparison} shows a more intuitive visualization for the comparison with~\cite{varol17_surreal},~\cite{bodynet2018},~\cite{laina2016deeper} and ~\cite{kovesi2005shapelets}. At the first row we show the heatmaps for depth errors. The method in SURREAL~\cite{varol17_surreal} produces incorrect human body segmentation, which leads to large errors at the boundary. The BodyNet~\cite{bodynet2018} has significant quantization errors due to the coarse volumetric representation. ~\cite{laina2016deeper} generates very rough depth maps with large structure error because of lacking intermediate supervision of 3D joints and segmentation. The result of ~\cite{kovesi2005shapelets} shows it can not handle depth discontinous cases such as when putting hands in front of the torso.

\subsection{Ablation Studies}
\label{sec:ablation}
In this section, we verify the effectiveness of the individual components of our method. To this end, we trained another 5 networks in the following settings and compared their results with ours.

\noindent \textbf{Without Skeleton and Segmentation Cues}: We discard Skeleton-Net and Segmentation-Net and only feed RGB image to Depth-Net to predict human body depth while the other conditions keep the same.

\noindent \textbf{Without Depth Separation}: We replace the two-branch architecture of the Depth-Net with only one branch. We train this network for the same epochs with the Huber loss defined as: 
\begin{equation}
    L = H(D_{pred} - D_{gt},  \alpha_{4}).
     \label{equ:setting1}
\end{equation}
where $\alpha_{4}$ is set as 0.20 meters in this setting.

\noindent \textbf{Only Stage 1 Training}: We keep the two-branches architecture and trained it only on stage 1 for the same total epochs. 

\noindent \textbf{Only Stage 2 Training}: The network is the same, while we train it directly on stage 2 without well-initialized weight of the base and detail branches.

\noindent \textbf{Huber Loss on Composed Shape}:  We follow the two stages training strategy on the same network but use Huber loss instead of truncated L1 loss to define the composed Loss in stage 2:
\begin{equation}
\begin{aligned}
    L_{composed} = H(D_{base} + D_{detail} - D_{gt}, \alpha_{5}).
\end{aligned}\label{equ:setting3}
\end{equation}
where $\alpha_{5}$ is 0.20m in this setting.

\begin{figure}
\begin{center}
\includegraphics[width=1\linewidth]{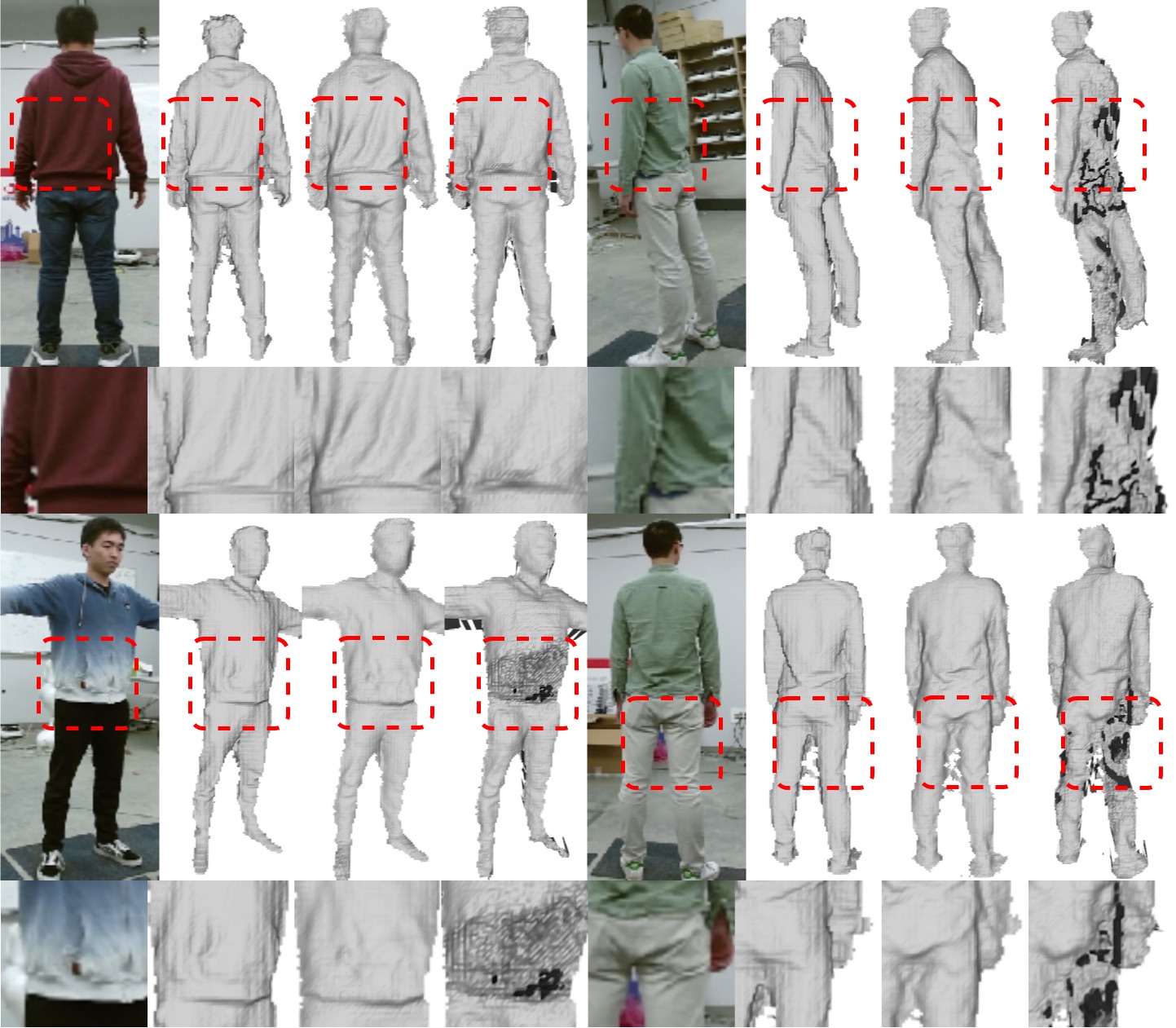}
\end{center}
\caption{Comparison of our proposed method and `W/o Skeleton and Segmentation Cues'. From left to right, they are the image, ground truth and the results from our method and the setting without Segmentation-Net and Skeleton-Net cues. It is clear that without high-level information to guide the depth estimation, the result might have large shape errors.}
\label{fig:comparison_no_clue}
\end{figure}

\begin{figure}
\begin{center}
\includegraphics[width=1\linewidth]{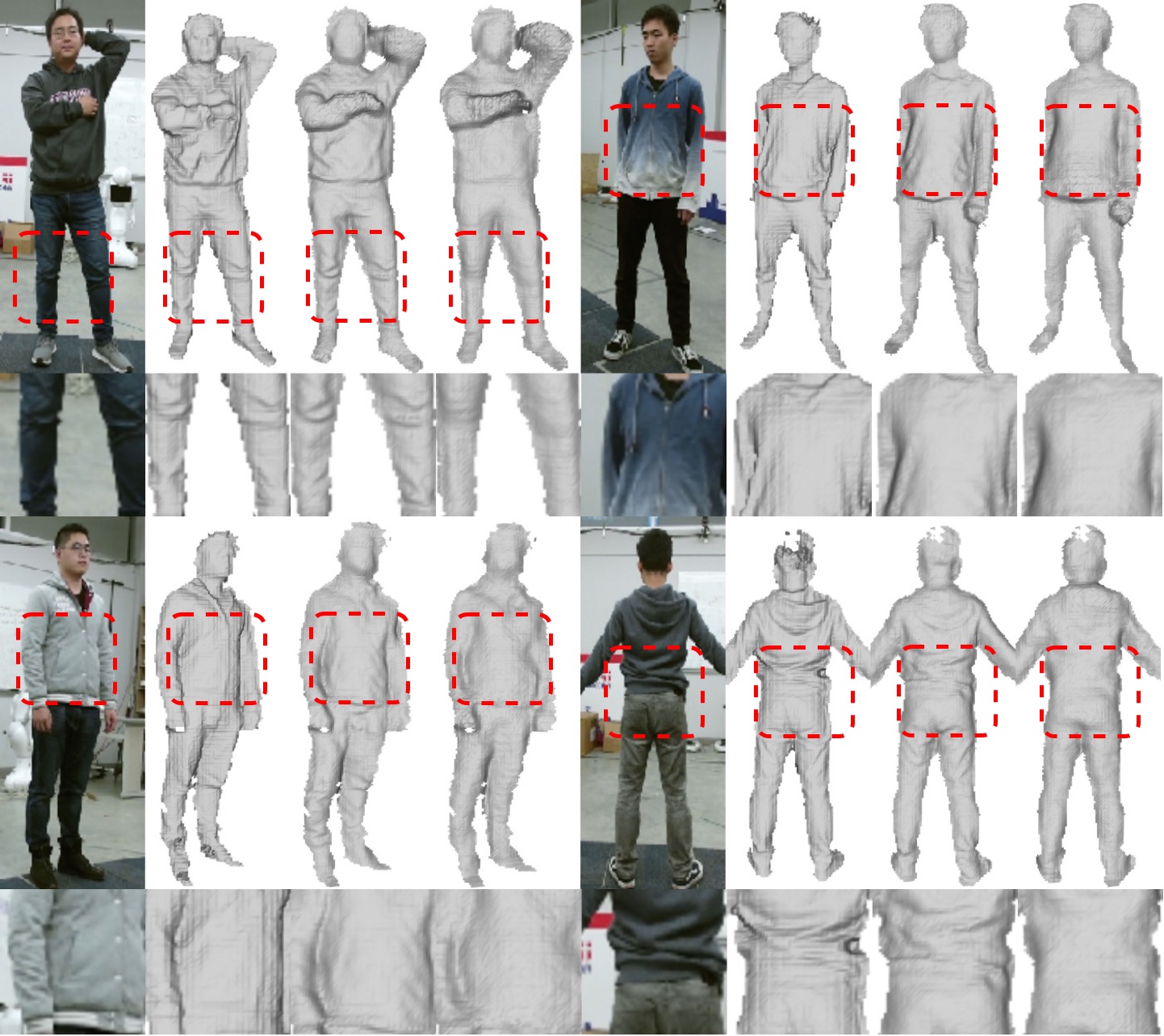}
\end{center}
\caption{Comparison of our proposed method and `No Depth Separation'. From left to right, they are the image, ground truth and the result from our method and the setting with only one depth branch. We can see that the results without a two-branch architecture are rough and do not have many geometry details.}
\label{fig:comparison_no_separation}
\end{figure}

\begin{figure}
\begin{center}
\includegraphics[width=1\linewidth]{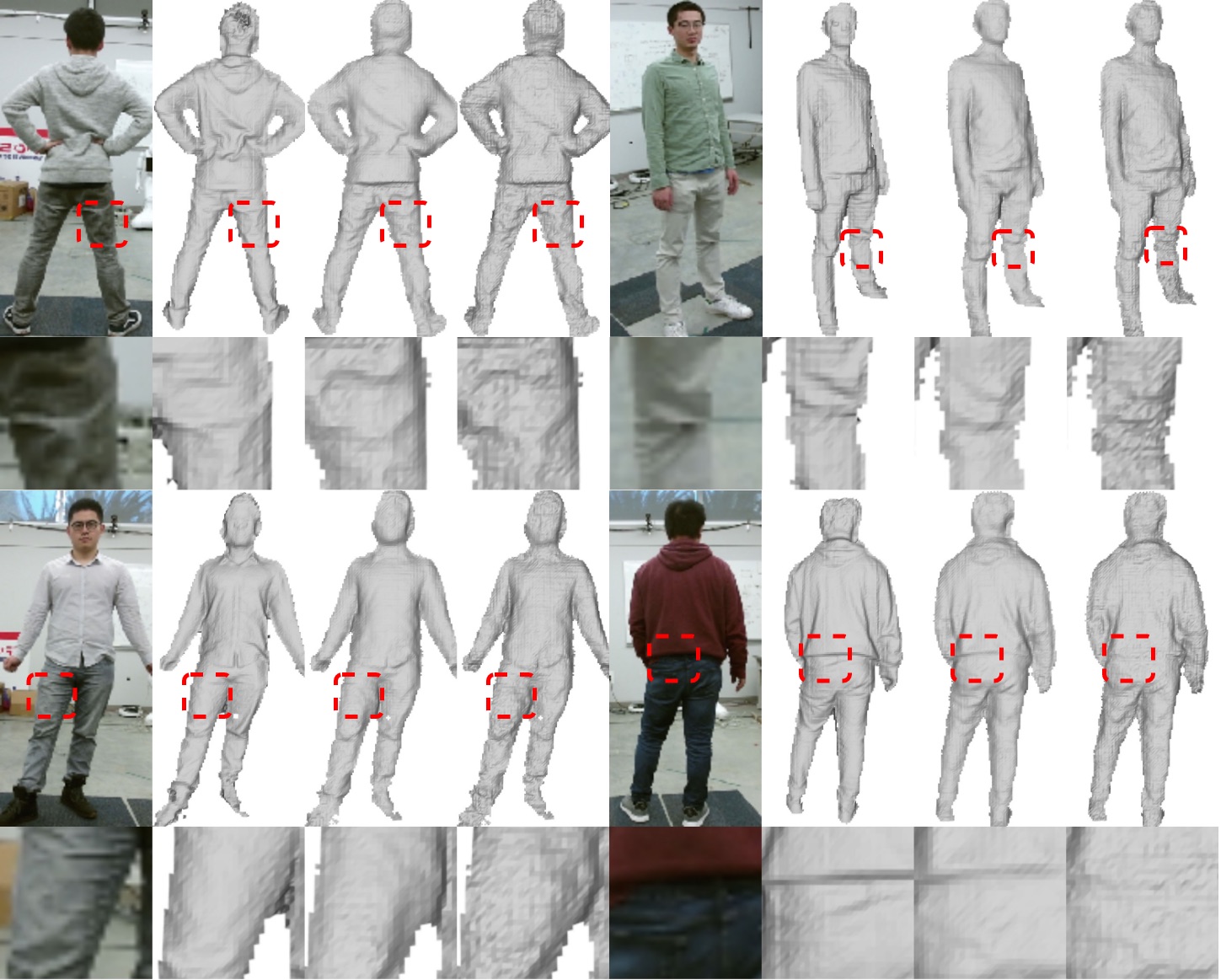}
\end{center}
\caption{Comparison of our proposed method and `Only Stage 1 Training'.From left to right, they are the image, ground truth and the result from our method and the setting with only stage 1 training. From the surface we can see the results without stage 2 will generate wrong wrinkles on the clothes.}
\label{fig:comparison_no_stage2}
\end{figure}

\begin{figure}
\begin{center}
\includegraphics[width=1\linewidth]{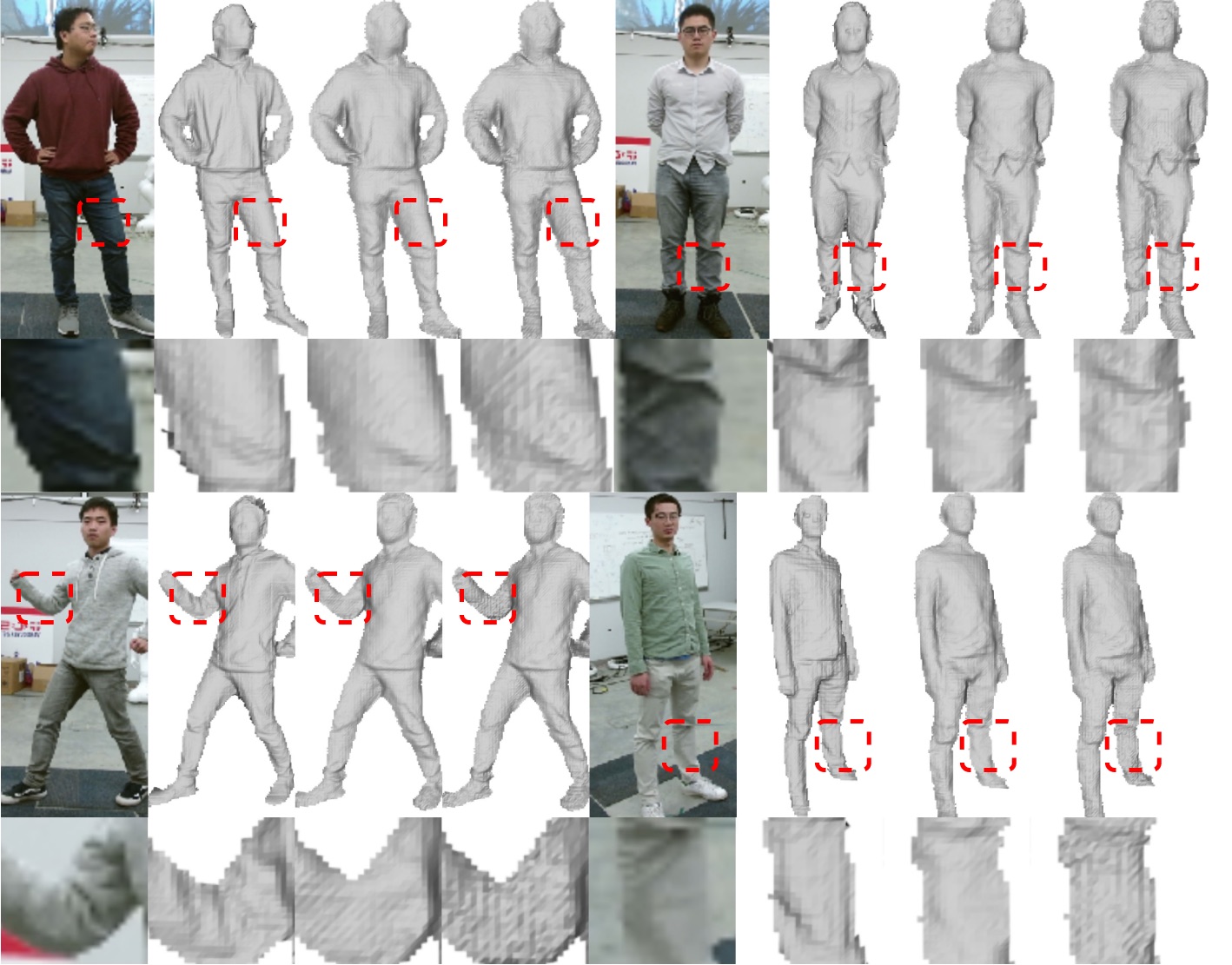}
\end{center}
\caption{Comparison of our proposed method and `Huber Loss on Composed Shape'. From left to right, they are the image, ground truth and the result from our method and the setting with Huber loss on the composed depth.  We can see the results without using our truncated L1 loss are unstable and not smooth enough.}
\label{fig:comparison_no_trucatedloss}
    \vspace{-4mm}
\end{figure}

\begin{figure}
\begin{center}
\includegraphics[width=1\linewidth]{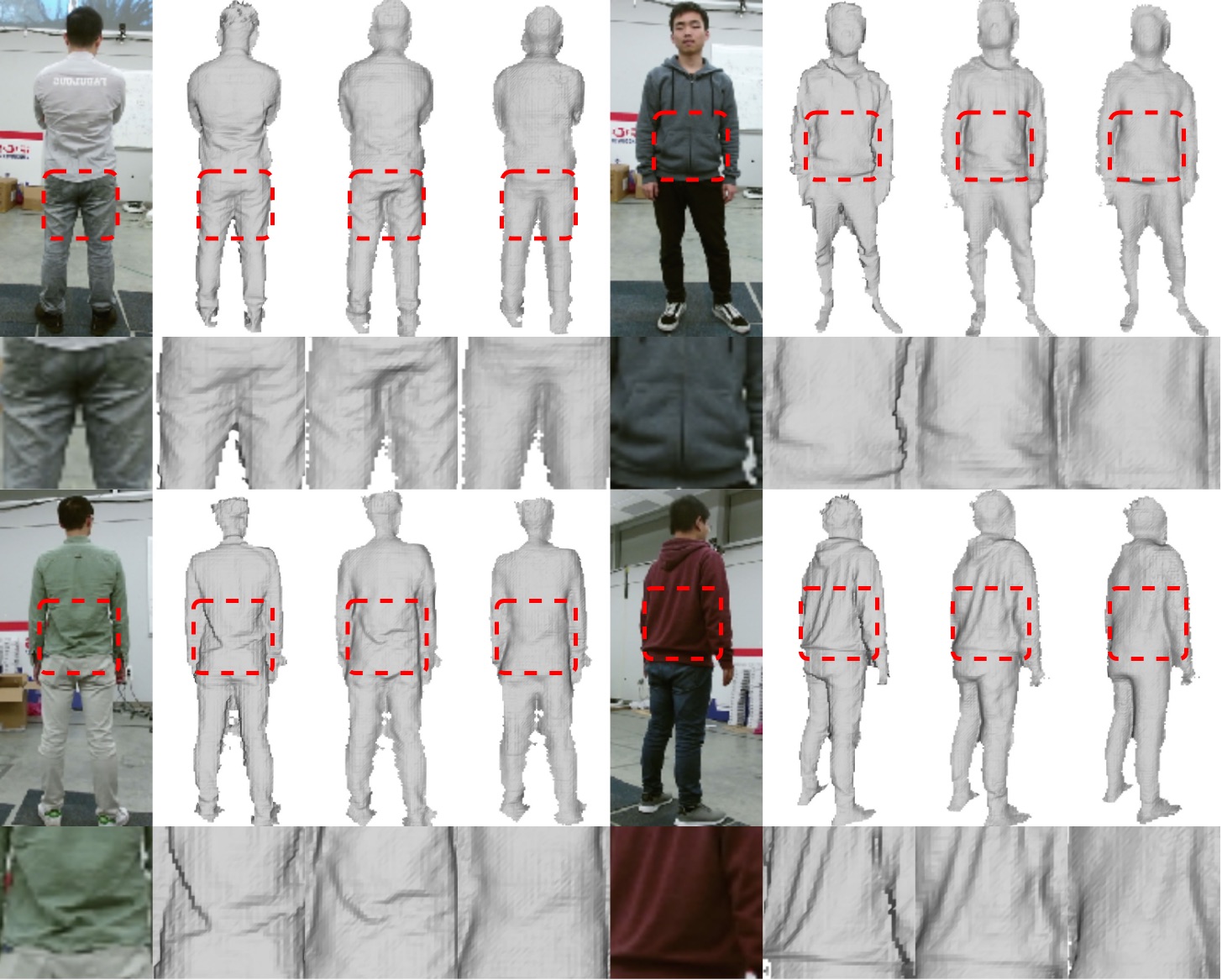}
\end{center}
\caption{Comparison of our proposed method and `Only Stage 2 Training'. From left to right, they are the image, ground truth and the result from our method and the setting with only stage 1 training. By zooming in the results we can see the setting without stage 1 lose majority of geometry details.}
\label{fig:comparison_no_stage1}
\end{figure}

We tested the five different settings mentioned above.
Figure~\ref{fig:comparison_no_clue},~\ref{fig:comparison_no_separation},~\ref{fig:comparison_no_stage2},~\ref{fig:comparison_no_trucatedloss}~\ref{fig:comparison_no_stage1}, show some qualitative comparisons of the results from these different settings. Specifically, Figure~\ref{fig:comparison_no_clue} shows that in the setting without Segmentation-Net and Skeleton-Net, the Depth-Net will lose the high-level human body information such as 3D joints and body part segmentation, hence the results show some structural issues, like broken meshes on some examples. Figure~\ref{fig:comparison_no_separation} clearly demonstrates that the network without a two-branch architecture is not able to recover small-scale geometry details. From the results of Figure~\ref{fig:comparison_no_stage2} and Figure~\ref{fig:comparison_no_trucatedloss} we can see that the recovered surface under these two settings are very coarse. Because without using truncated L1 loss which clips the composed error in stage 2 to improve the consistency of two branches, the large layout error may overwhelm the detail error and leads to unstable results from two branches. Figure~\ref{fig:comparison_no_stage1} shows without  stage 1 guiding two branches focusing on their target distribution, the detail branch is not working on recovering the small wrinkle specifically. In summary, it is clear that our method produces the best shape details, main layout and smooth surface, which demonstrates the effectiveness of separating the base shape and detail shape and two-stage training with truncated L1 loss on the composed shape. 

\begin{table}[]
\begin{tabular}{|l|lll|l|}
\hline
\multicolumn{1}{|c|}{Methods} & \multicolumn{3}{|c|}{Accuracy}                                                                           & MAE \\ \cline{2-4}
\multicolumn{1}{|c|}{}                        & \multicolumn{1}{c}{1.25cm}           & \multicolumn{1}{c}{2.5cm}           & \multicolumn{1}{c|}{5.0cm} &                      \\ \hline
Ours                             & \multicolumn{1}{l|}{\textbf{29.24}} & \multicolumn{1}{l|}{\textbf{50.93}} & \textbf{75.52}             & \textbf{3.282}       \\ \hline
W/o seg\&skeleton                                       & \multicolumn{1}{l|}{25.74}          & \multicolumn{1}{l|}{46.19}          & 71.04                      & 4.382                \\ \hline
W/o separation                              & \multicolumn{1}{l|}{28.00}          & \multicolumn{1}{l|}{49.42}          & 72.97                      & 3.480                \\ \hline
Only stage 1                            & \multicolumn{1}{l|}{26.64}          & \multicolumn{1}{l|}{48.14}          & 72.61                      & 3.592               \\ \hline
Only stage 2                                 & \multicolumn{1}{l|}{27.89}          & \multicolumn{1}{l|}{50.31}          & 74.87                      & 3.332                \\ \hline
W/o truncated loss                                       & \multicolumn{1}{l|}{28.03}          & \multicolumn{1}{l|}{49.84}          & 74.23                      & 3.410                \\ \hline
\end{tabular}
\\[1pt]
\caption{Performance of depth prediction of our method and other five settings in Section~\ref{sec:ablation}}
\label{tab:AUC_4settings}
\vspace{-2mm}
\end{table}


Table~\ref{tab:AUC_4settings} further provides the quantitative comparison of these settings on our testing data with fused ground truth depth~\cite{InfiniTAM_ISMAR_2015}. Note that compared results are the composed shape without refinement. The proposed method consistently outperforms the other settings.




\subsection{Qualitative Results}
\begin{figure}
\center
 \includegraphics[width= 0.9 \linewidth]{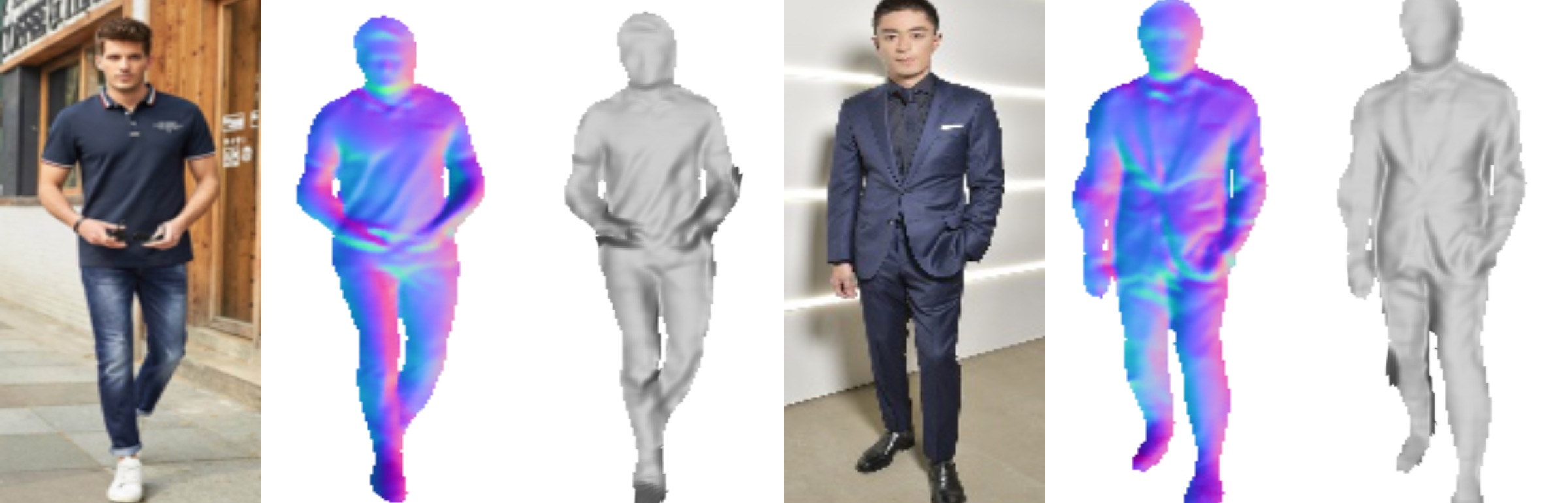}\\
 \includegraphics[width= 0.9 \linewidth]{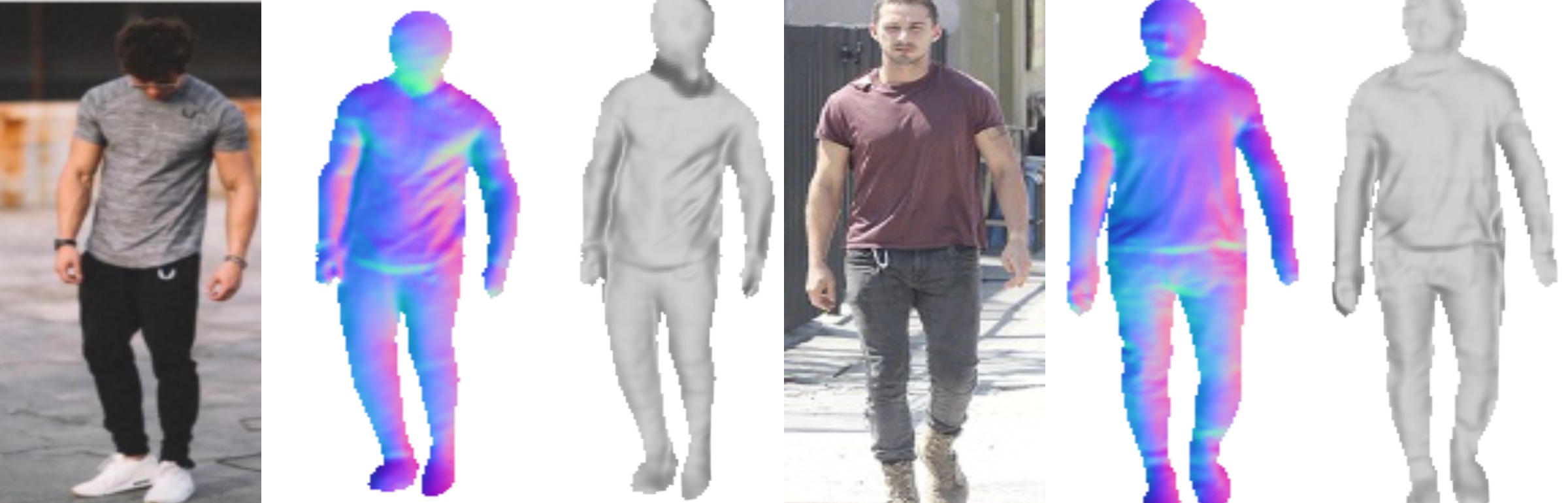}\\
 \includegraphics[width= 0.9 \linewidth]{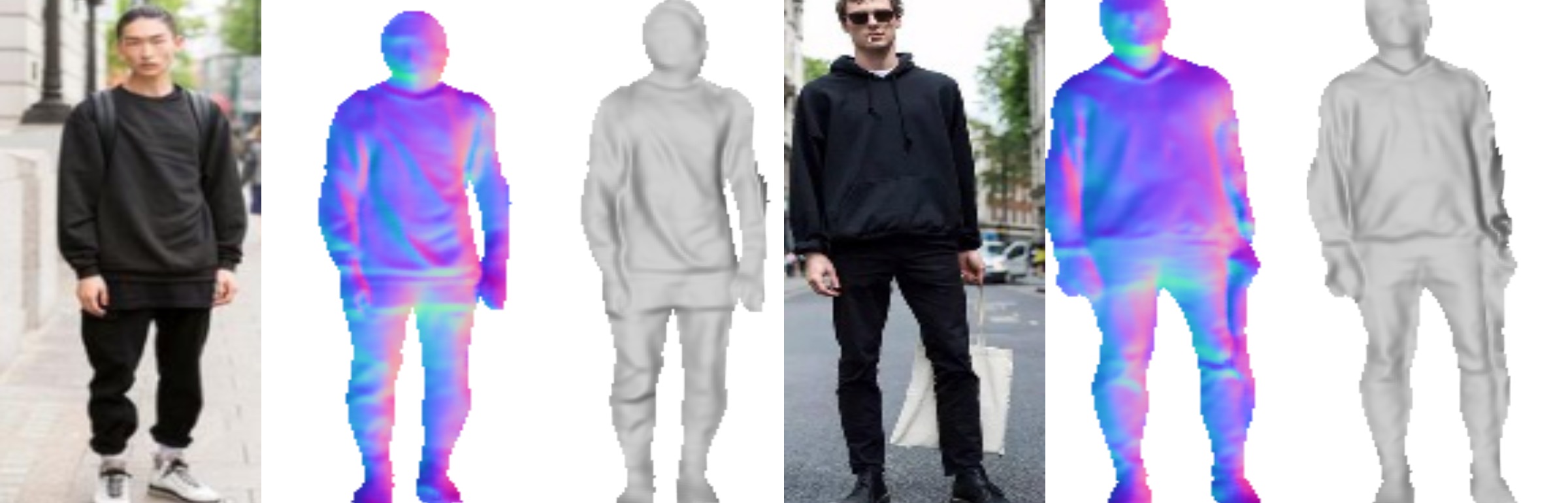}
\caption{Some results on unconstrained online images. From left to right, for each example, we show the input image, estimated surface normal and final result.}
\label{fig:onlinedata}
    \vspace{-4mm}
\end{figure}

To demonstrate our network can be generalized to unconstrained data, Figure~\ref{fig:onlinedata} shows our results on some unconstrained Internet images. Our method also successfully recovers certain shape details on these images. We further visualize the estimated surface normal map, which encodes the cloth wrinkles. 

In our demo video, we demonstrate the performance of our method on some video clips, which are processed in a frame-by-frame fashion. The result shows that our method can even generate temporally coherent results without explicitly modeling it. 



\section{Conclusion}
This paper proposes a neural network to estimate a detailed depth map for the human body in a single input RGB image. The recovered result can capture fine cloth wrinkles and produce temporally coherent depths for video inputs. It might be used in visualization applications such as the Microsoft Holoportation. This result is achieved by separating and estimating the base shape and detail shape respectively with a novel truncated L1 loss. We also introduce a novel parameter free shape refinement layer to further improve the final result with surface normals. Quantitative evaluation on lab data and qualitative examples on unconstrained Internet data demonstrate the success of the proposed method. 

{\small
\bibliographystyle{ieee_fullname}
\bibliography{egbib}
}

\end{document}